\documentclass[nohyperref]{article}

\usepackage{microtype}
\usepackage{graphicx}
\usepackage{subcaption}
\usepackage{booktabs} 
\usepackage{listings}

\usepackage{hyperref}
\usepackage{stix}
\usepackage{multirow}
\usepackage{array,multirow}
 \usepackage{booktabs}
 \usepackage{multirow}

\usepackage{natbib} 
\usepackage{mathtools} 
\usepackage{booktabs} 
\usepackage{tikz} 

\usepackage[accepted]{icml2022}

\usepackage{amsmath}
\usepackage{amssymb}
\usepackage{mathtools}
\usepackage{amsthm}

\usepackage[capitalize,noabbrev]{cleveref}
\theoremstyle{plain}

\theoremstyle{definition}

\theoremstyle{remark}

\usepackage[textsize=tiny]{todonotes}

\icmltitlerunning{Selection Bias Induced Spurious Correlations in Large Language Models}

\begin{document}

\twocolumn[
\icmltitle{Selection Bias Induced Spurious Correlations in Large Language Models}



\icmlsetsymbol{equal}{*}

\begin{icmlauthorlist}
\icmlauthor{Emily McMilin}{id}

\end{icmlauthorlist}

\icmlaffiliation{id}{Independent Researcher}

\icmlcorrespondingauthor{Emily McMilin}{emcmilin@cs.stanford.edu}

\icmlkeywords{Machine Learning, ICML, Spurious Correlations, Causal Inference}

\vskip 0.3in
]



\printAffiliationsAndNotice{}  

\begin{abstract}
In this work we show how large language models (LLMs) can learn statistical dependencies between otherwise unconditionally independent variables due to dataset selection bias. To demonstrate the effect, we developed a masked gender task that can be applied to BERT-family models to reveal spurious correlations between predicted gender pronouns and a variety of seemingly gender-neutral variables like date and location, on pre-trained (unmodified) BERT and RoBERTa large models. Finally, we provide an online demo, inviting readers to experiment further.

\end{abstract}

\section{Introduction}\label{intro}

For generalization to real-world target domains, a learned model's training data would ideally be a randomized sampled subset of the data in the target domain. However, achieving such randomization is often impractical, resulting in many datasets exhibiting sampling bias \cite{heckmanSamplingBias}. Models trained on datasets with sampling bias are vulnerable to learning spurious associations from this bias. 

These spurious associations, often referred to as spurious correlations though the associations need not be linear, can reduce the predictive performance of the model in real-world domains. Specifically, although we desire the model to learn the conditional distribution: $P( Y | X )$, it has instead learned $P( Y | X, S\negthickspace=\negthickspace1 )$, where $S\negthickspace=\negthickspace1$ represents selection into the dataset \cite{pearlSelection12}.
 
This paper focuses on a specific type of selection bias in which two variables: $W\negthinspace$ and $G$, which are unconditionally independent in the real world ($G \!\perp\!\!\!\perp W$), become conditionally dependent within the dataset ($G \not \!\perp\!\!\!\perp  W | S$), due to the selection process, $S$.

We hypothesize that a wide range of spurious correlations can be traced back to this type of bias. To expose subtle spurious associations that have not yet been reported, we desire an underspecified learning task, in which there are multiple equally plausible predictions. In natural language processing, one well-researched underspecified task is that of gender pronoun prediction \cite{underspec}, in which a gender is predicted from gender-neutral features.

Undesirable and spurious associations between gender: $G$, and variables: $W^\prime\negthickspace$,  such as occupation \cite{websterGenderCorr20} and college major \cite{Rudinger18} have been found in LLMs.
Unfortunately, many of these spurious associations are in fact representative of undesirable gender inequities in our real world ($G \not \!\perp\!\!\!\perp  W^\prime$).

To show how selection bias can induce associations that do not match our real-world distributions of gender, we seek to demonstrate spurious associations between gender and real-world gender-neutral variables: $W\negthickspace$. 

In this paper we introduce and use the `masked gender task' to demonstrate previously unreported spurious correlations between gender pronouns and the following gender-neutral entities for $W\negthinspace$: time, location, and topics of interest, on unmodified pre-trained BERT large \cite{BERT} and RoBERTa large  \cite{RoBERTa} models.

\section{Data Generating Processes}\label{dgp}
A sample can only be selected into a dataset if it has \textit{access} to the sampling process. We use the term \textit{access} here, as it evokes processes of selection also experienced by human-beings, which we will further motivate when describing the data generating process. 

To understand the selection bias intrinsic to a dataset, one must consider the dataset's data generating process. 
Datasets do not generally admit their generating process, but rather it must be discovered via auxiliary methods such as applying domain knowledge or causal discovery methods. We use intuition and domain knowledge to describe what we hope are plausible data generating processes below.

\subsection{Datasets}

We ground this discussion with two example datasets, selected as they are representative of data sources used to pre-train LLMs. 

Specifically, we use the Wikipedia Biography (Wiki-Bio) dataset \cite{wikibio}, composed of about 730,000 biographies from English Wikipedia, with which we finetuned models on the roughly 105,000 entries that contained birth \textit{date} and birth \textit{place} data, as well as at least one instance of an explicitly gendered word, from the list in \cref{tab:non-gender-neutral}\footnote{Further selection bias is introduced here during the filtering of ineligible entries, of unknown effect on finetuned models, which serve referential purposes for the pre-trained model's results.}.

And we use Reddit Webis-TLDR-17 (Reddit-TLDR) dataset \cite{reddit}, composed of content-summary pairs from Reddit from 2006-2016, with which we finetuned models on the roughly 320,000 entries that again contained at least one explicitly gendered word from \cref{tab:non-gender-neutral}. Both datasets are hosted on Hugging Face's Datasets Library.

\subsection{Selection Variables}

Recall we desire to demonstrate a learned statistical association between gender, $G$, and gender-neutral variables, $W\negthinspace$, that is driven by \textit{access} (now referred as $Z$) to the dataset sampling process. 
For our datasets, suitable instantiations for $W\negthinspace$ and $Z$ as related to $G$ are as follows. 

For Wiki-Bio: \textit{access} to resources has generally become less gender inequitable over time as \textit{date} ($W_0$) increases, but not evenly in every \textit{place} ($W_1$). Generally only those with \textit{access} to resources will achieve the level of notoriety necessary for an entry in Wikipedia.

For Reddit-TLDR: despite many \textit{subreddit} channels ($W\negthinspace$) having a focus on gender-neutral topics of interest, the style of moderation and community within a given subreddit may reduce gender-equal \textit{access} to participation in that \textit{subreddit}.

\subsection{Causal DAGs}

\begin{figure}
\begin{center}
  \begin{subfigure}[b]{0.48\columnwidth}
      \centering
      \includegraphics[width=\columnwidth]{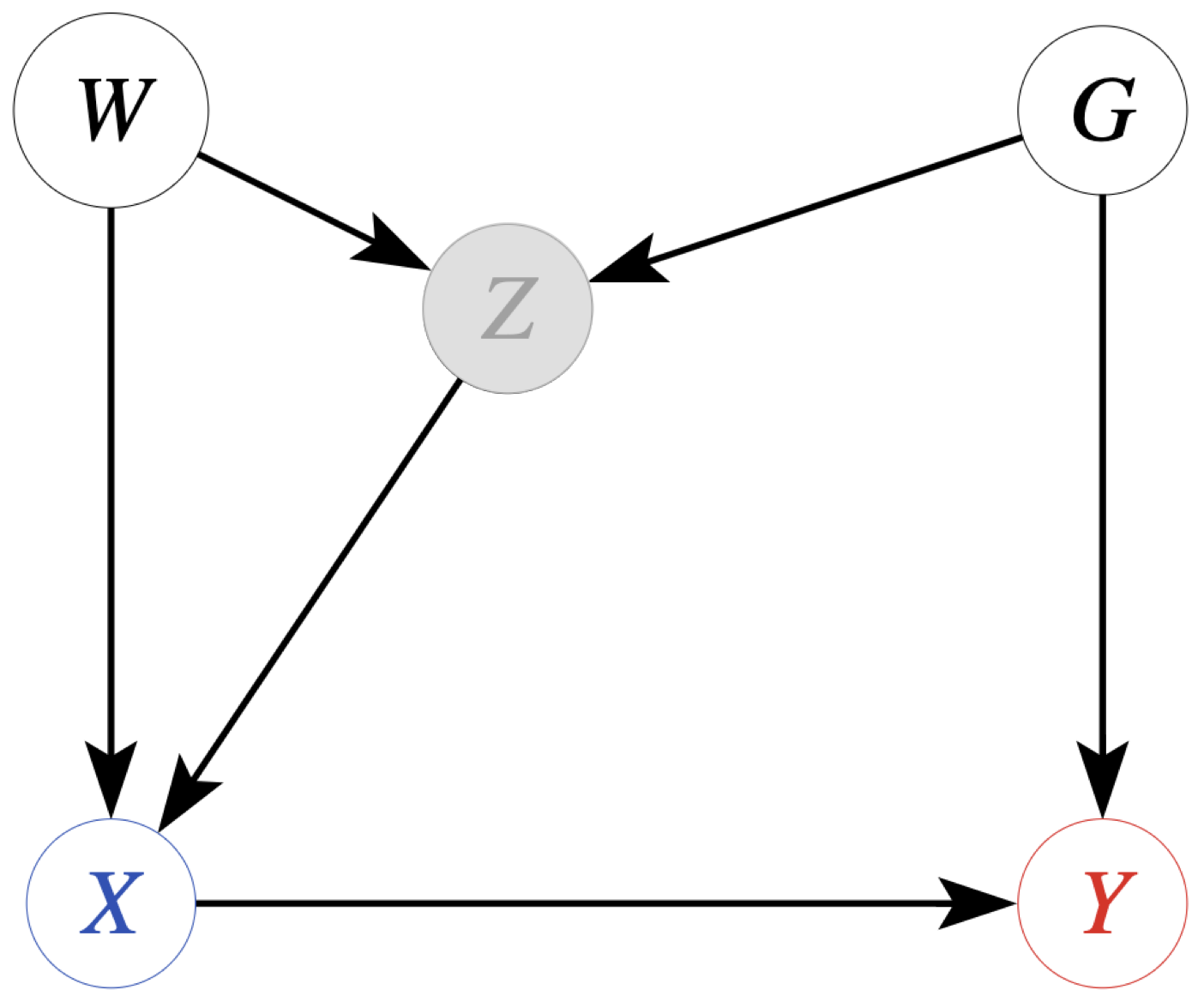}
      \caption{DAG including $G$: gender, that is later unobserved.}
      \label{fig-DAG-w-G}
  \end{subfigure}
  \hfill
  \begin{subfigure}[b]{0.48\columnwidth}
      \centering
      \includegraphics[width=\columnwidth]{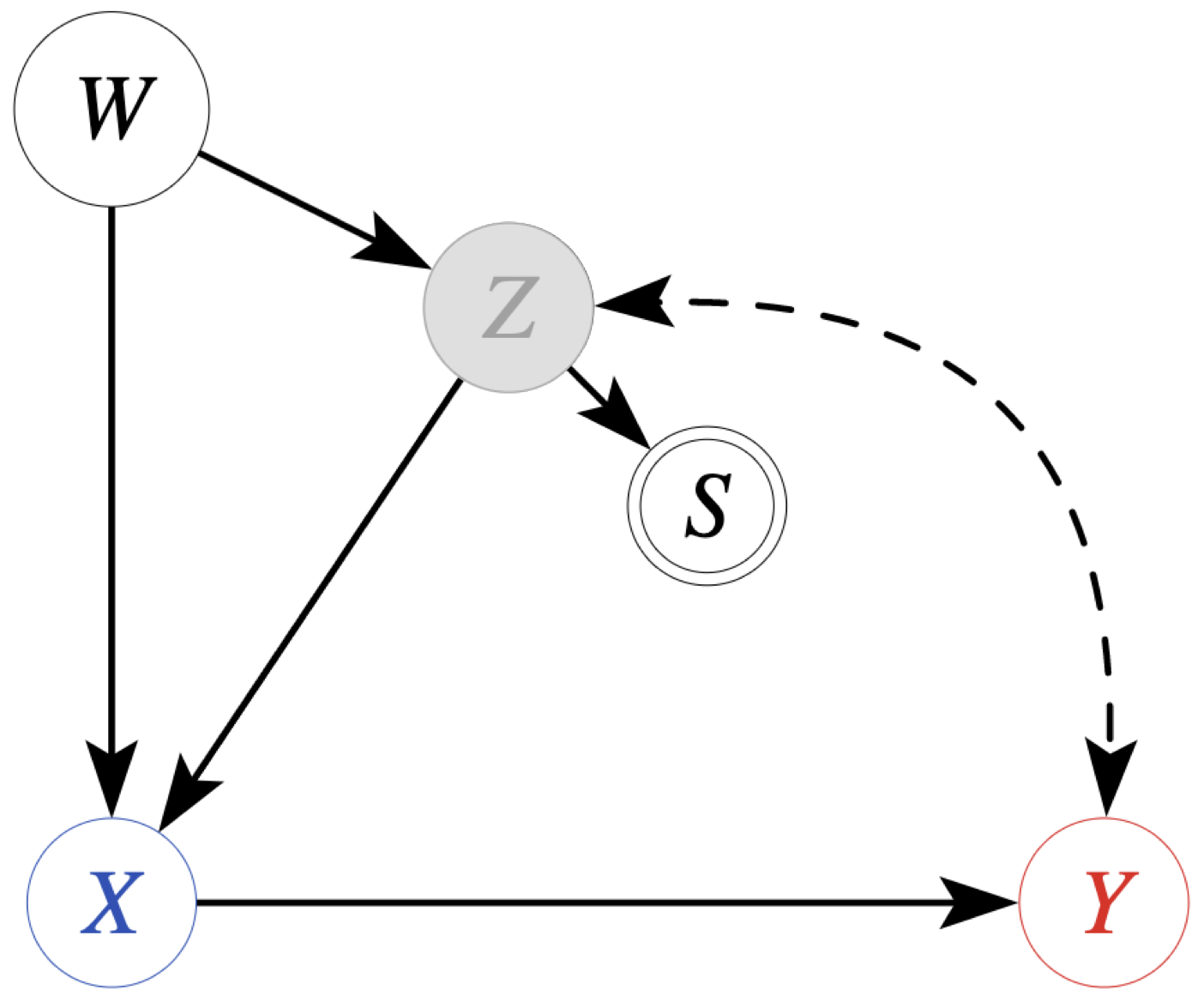}
      \caption{DAG including node, $S$, representing the dataset selection.}
      \label{fig-DAG-w-S-node}
  \end{subfigure}
  \label{fig-DAGs}
    \vskip -.15in
    \caption{Causal DAG representing the assumed data generating process for the Wiki-Bio and Reddit TLDR datasets. }
  \end{center}
  \vskip -0.3in

\end{figure}

The above written descriptions of variables and their causal relationships can be compactly represented in a causal directed acyclic graph (DAG), where nodes are variables and arrows are the direction of causation, as can be seen in \cref{fig-DAG-w-G} for our assumed data generating process for the Wiki-Bio and Reddit TLDR datasets.

From \cref{fig-DAG-w-G}, we see that $W\negthinspace$ and $G$ can cause one's \textit{access}, $Z$, to dataset selection, as discussed above.

At the bottom we see our dataset's features: $X$ for \textit{text}, and labels: $Y$ for \textit{pronouns}. We argue that despite the complex causal interactions between words in a sentence, the \textit{text} are more likely to cause the \textit{pronouns}, rather than vice versa.\footnote{For example, if the subject is a famous doctor and the object is her wealthy father, these context words will determine which person is being referred to, and thus which gendered-pronoun to use.} 

Further we assume that $Z$ and $W$ have an effect on one's life and thus the \textit{text} written about them or by them. And clearly $G$ does cause the gender \textit{pronouns}, $Y\negthinspace$. However, because the goal of the masked gender task is to mask out explicitly gendered words, we'd argue that $G$ is not a direct cause of the \textit{text}, $X$. Finally, note $Z$ is grayed out, as it is not explicitly recorded in the dataset.

\section{Selection Bias}
To explain how selection bias can cause two unconditionally independent variables to become dependent, we revisit the causal DAG representing the data generating processes in \cref{fig-DAG-w-G}. Causal DAGs are associated with a set of structural equations that together compose a structural causal model \cite{Pearl09}. For the \textit{access} variable, the structural equation is \(Z\negthickspace:= f_z(W, G, U_z)\label{eq-z-scm}\), where $U_z$ is the exogenous noise of the $Z$ variable, and again $W\negthinspace$ are the variables \textit{date} and \textit{place} for Wiki-Bio, and \textit{subreddit} for Reddit TLDR, and $G$ is gender. Although $W\negthinspace$ and $G$ are independent in \cref{fig-DAG-w-G}, for all but trivial special cases, they will become statistically associated in the equation $f_z$ for $Z$. 

\subsection{Conditioning on a Collider}
Thus any model that conditions on the variable $Z$, will introduce this spurious association between $W\negthinspace$ and $G$, known as collider bias \cite{Pearl09}. Excluding $Z$ from the predictive model seems trivial, especially because $Z$ is not directly observed in the dataset. However, recall that $Z$ is a cause of the selection process, depicted in the selection diagram, \cref{fig-DAG-w-S-node}, where the selection node, $S$, can only take values  \(S\negthickspace=\negthickspace1\) for a sample selected into the dataset and \(S\negthickspace=\negthickspace0\) otherwise \cite{pearlSelection12}. 

Recall from \cref{intro} that during dataset formation, we implicitly condition on $S\negthickspace=\negthickspace1$, as only selected samples appear in the dataset. Conditioning on a descendant of $Z$, induces the collider bias relationship, as if we had conditioned on $Z$ directly. We are thus inducing the latent structural equation for $f_z$ into the dataset, from which the statistical association between $W\negthinspace$ and $G$ can be learned by our models.

\subsection{Selection Bias Recoverability}\label{recovery}
\cref{fig-DAG-w-S-node} is a modified version of our original causal DAG, which satisfies the requirements of the masked gender task, specifically we have obscured $G$ and replaced it with double headed arrows to represent an unobserved common cause of both $Z$ and $Y$. 

Although the act of obscuring gender for gender pronoun prediction may seem contrived, we argue that LLMs are often in similar circumstances, for example whenever a prompt, dialog, translation or classification task has not been provided gender features, yet predictions about gender are required.

Structural causal models very similar to that in \cref{fig-DAG-w-S-node} have been described in practice in \cite{knox2020}, and proven in \cite{selectionrecover}, to be not `recoverable'. Formally, because we are unable to d-separate the selection mechanism from the label: $(Y\not\!\perp\!\!\!\perp  S | X)$, the conditional distribution of \(P(Y|X)\) cannot be determined without further assumptions or additional data about target populations \cite{BareinboimTian2015}.

\section{Masked Gender Task}

This lack of recoverability for \(P(Y|X)\) is consistent with our goals of underspecification for the masked gender task.

\subsection{Inference Test Texts}\label{sec:input-texts}
Revisiting \cref{fig-DAG-w-S-node}, to maintain underspecification we now require gender-neutral text values for the input \textit{text}, $X\negthinspace$, and the \textit{date}, \textit{place}, and \textit{subreddit} variables, $W\negthinspace$.
In addition to gender-neutral, for $X$ we desire extremely simplistic texts, to avoid inducing unrelated spurious correlations. We were unable to find such a benchmark dataset, so we used the heuristic described in \cref{tab:input-text} to generate over 700 input texts for each of the three $W\negthinspace$ categories.

\begin{table*}[t]
\caption{Heuristic for creating gender-neutral input texts for the masked gender task for each $W\negthinspace$ variable category, and example rendered text. For  \texttt{verb} we used past, present and future tenses of the verb \textit{to be}: {[}"was","is","will be"{]}, and for \texttt{life\_stage} we used proper and colloquial terms for a range of life stages: {[}"a child", "a kid", "an adolescent", "a teenager", "an adult", "all grown up"{]}. We didn't include any \texttt{life\_stage} past adulthood, as there are not equal gender ratios of elderly men to women, in many locations. Finally for the text versions of \texttt{w}, we used a spectrum of values defined in \cref{sec:x-axis}.}
\begin{center}
\begin{small}
\begin{tabular}{lll}
\toprule

$W\negthinspace$ Category                                                    & Python f-string                                              & Example text                                                             \\
\midrule
\midrule
\multicolumn{1}{c}{\multirow{2}{*}{Date \& Place}} & \texttt{`f"{[}MASK{]} \{verb\} \{life\_stage\} in \{w\}."'}           & `{[}MASK{]} was a teenager, in 1953.'                                        \\
\multicolumn{1}{c}{}                                 & \texttt{`f"In \{w\}, {[}MASK{]} \{verb\} \{life\_stage\}."'}        & `In Mali, {[}MASK{]} will be an adult .'                          \\
\midrule
Subreddit                                            & \texttt{`f"{[}MASK{]} \{verb\} \{life\_stage\}. \{w\}."'} & `{[}MASK{]} is a kid. gifs.' \\

\bottomrule
\end{tabular}
\label{tab:input-text}
\end{small}
\end{center}
\vskip -0.15in
\end{table*}

\subsection{$W\negthinspace$  x-axis Values}\label{sec:x-axis}
Remaining undefined in the heuristic for input texts in \cref{tab:input-text}, is the text values to use for $W\negthinspace$. These $W\negthinspace$ values will also serve as our x-axis in the coming plots. We require values that are gender-neutral in the real world, yet are hypothesized, due to the selection bias process, to be a spectrum of gender-inequitable values in the dataset.

For $W\negthinspace$ as \textit{date}, it's easy to just use time itself, as over time women have become more likely to be recorded into historical documents reflected in Wikipedia, so we pick years ranging from 1801 - 2011.
For $W\negthinspace$ as \textit{place}, we use the bottom and top 10 World Economic Forum Global Gender Gap ranked countries (see \ref{place-list}).
And for $W\negthinspace$ as \textit{subreddit}, we use \textit{subreddit} name ordered by subreddits channels that have an increasingly larger percentage of self-reported female commenters, with a minimum size of 400,000 commenters overall (see \ref{subreddit-list}). 

The original premise for the $W\negthinspace$ variable was to use categories that are gender-neutral in the real world, but not necessarily so in the dataset. To achieve this, we considered filtering the \textit{subreddit} list to only topics deemed gender-neutral, however this subjective process invited too much cherry picking on our behalf. Thus, for the \textit{subreddit} (and \textit{place}) values, we copied the referenced lists verbatim from their sources.

To help disambiguate the role of non-gender-neutral subreddit topic names, from the role of \textit{access} based selection bias, in contributing to a correlation between $W\negthinspace$ and $G$, we tested on one pre-trained model likely exposed to the selection bias effect during pre-training, and one that was likely not exposed. Specifically, RoBERTa was trained with the same data sources as BERT, plus additional data sources including OpenWebText\footnote{Although OpenWebText does not explicitly include Reddit data, because it is composed of scraped web content from URLs shared on Reddit that received at least 3 upvotes \cite{RoBERTa}, we conjecture subreddit topic names would appear in the context of Reddit in this dataset.} \cite{OpenWebText}. Thus, we'd expect RoBERTa to exhibit a stronger \textit{subreddit} to \textit{gender} correlation, than that of BERT, due to the role of subreddit \textit{access} selection bias during the pre-training of RoBERTa.

\subsection{Pre-trained BERT-like models}

For testing the masked gender task on pre-trained models, we selected BERT large and RoBERT large, as explained in \cref{sec:x-axis}, using the default weights hosted for the models on Hugging Face.

We are able to test the pre-trained LLMs without any modification to the models, as the masked gender task is simply a special case of the masked language modeling (MLM) task, with which all these models were pre-trained. 

Rather than random masking, the masked gender task masks only explicitly gendered words (listed in \cref{tab:non-gender-neutral}). During LLM pre-training, the MLM prediction is a softmax over the entire tokenizer's vocabulary. The masked gender task sums the gendered portion (as listed in \cref{tab:non-gender-neutral}) of that probability mass from the top five predicted words.

\subsection{Finetuned Models}

We also finetune BERT-like models using a similar masked gender task. The difference being that for our finetuning task, the prediction outcome is binary (as opposed to the entire tokenizer's vocabulary), largely for run-time expediency. We elected to finetune the models with data sources similar to those in their pre-training, so we selected BERT for the Wiki-Bio dataset and RoBERTa for the Reddit TLDR dataset. 

For the Wiki-Bio dataset, we finetune three BERT base models: 1) with birth \textit{date} metadata, 2) birth \textit{place} metadata, and 3) with no extra metadata, prepended to each training sample. In the case of the Reddit TLDR dataset we finetune two RoBERTa base models: 1) with \textit{subreddit} metadata and 2) with no extra metadata, prepended to each training sample.

 As these models are finetuned with a single dataset (Wiki-Bio or Reddit TLDR) for the single task of gendered-word prediction, we'd expect them to serve as an upper limit for the magnitude of spurious correlations a model could exhibit for the masked gender task. In particular, the models conditioned with the textual metadata values for $W\negthinspace$ at train time are expected to learn the strongest relationship between $W\negthinspace$ and $G$.

\section{Results}

In this section we share the results of the masked gender task, tested on pre-trained BERT and RoBERTa large, as well as our finetuned models which can serve as a rough upper limit for the magnitude of expected spurious correlations.

\begin{figure*}[t]
\vskip -0.05in
  \centering
  \begin{subfigure}[b]{0.32\textwidth}
      \centering
	{\includegraphics[width=\textwidth]{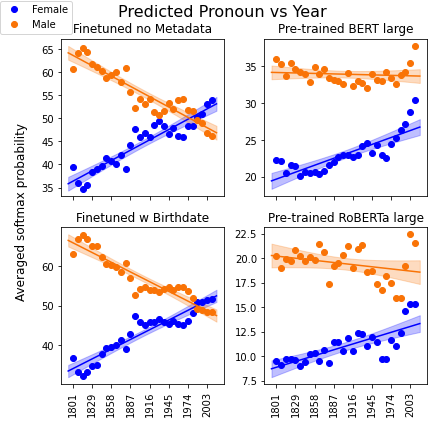}}
  	\caption{$W\negthinspace$ as \textit{date}, each dot is an average of 36 softmax probabilities for the mask in input text like “{[}MASK{]} will be an adult in 1945.”.}
  	\label{date-multiplot}
  \end{subfigure}
  \hfill
   \begin{subfigure}[b]{0.32\textwidth}
      \centering
	{\includegraphics[width=\textwidth]{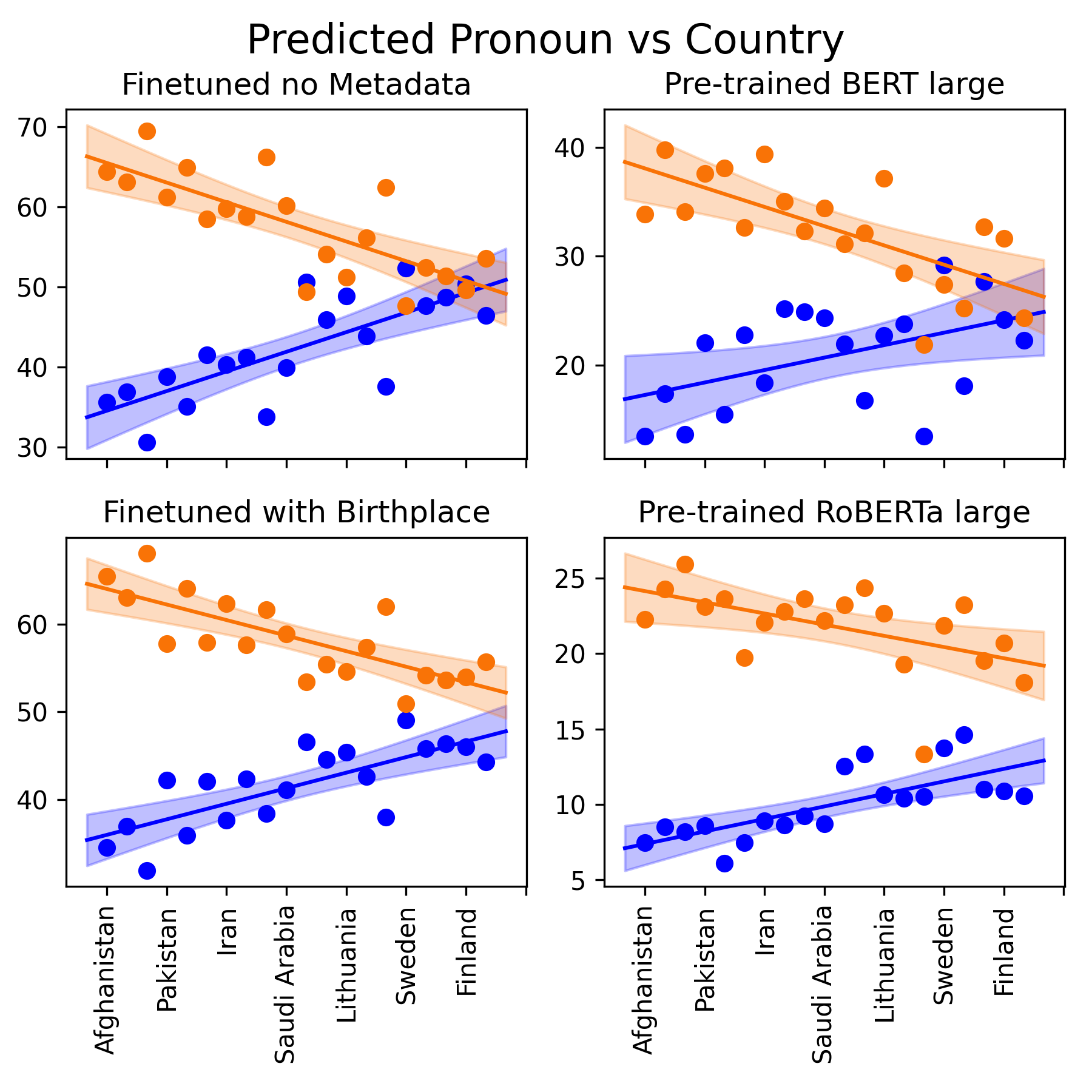}}
	 \caption{$W\negthinspace$ as \textit{place}, each dot is an average of 36 softmax probabilities for the mask in  input text like “In Iran, {[}MASK{]} was a teenager.”.}
  	  \label{place-multiplot}
  \end{subfigure}
  \hfill
     \begin{subfigure}[b]{0.33\textwidth}
      \centering
	{\includegraphics[width=\textwidth]{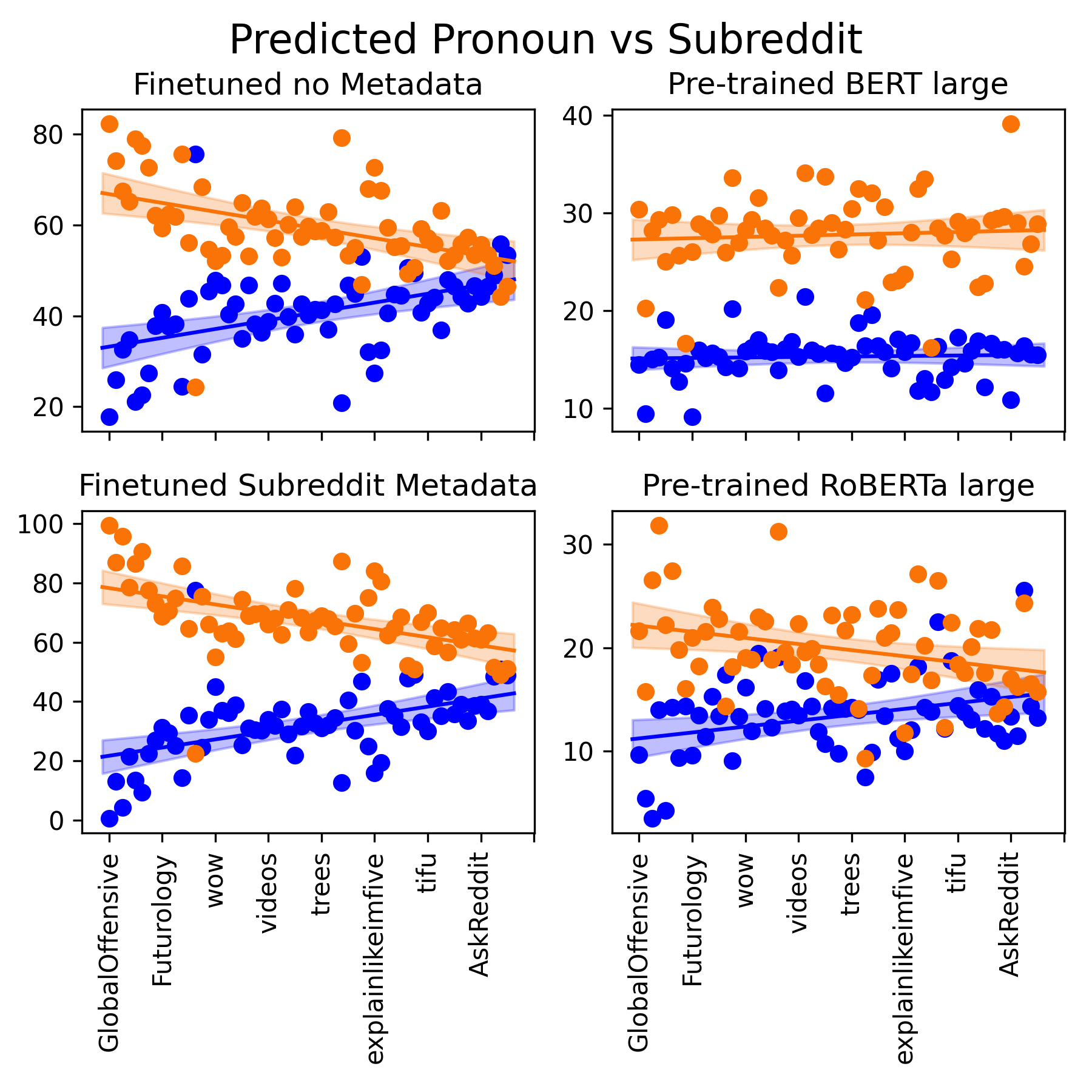}}
  	\caption{$W\negthinspace$ as \textit{subreddit}, each dot is an average of 18 softmax probabilities for the mask in  input text like “{[}MASK{]} is a kid. Futurology”.}
  	\label{subreddit-multiplot}
  \end{subfigure}
    \vskip -0.15in
  \caption{Averaged softmax percentages for predicted gender pronouns vs a range of $W\negthinspace$ values as described in \cref{sec:x-axis}, for gender-neutral input texts described in \cref{tab:input-text}. Shaded regions show the 95\% confidence interval for a 1st degree linear fit between $W\negthinspace$ and $G$. For all plots we expect a correlation between the x-axis  and the predictions, except for Pre-trained BERT large's predictions vs subreddit. }
\label{all-multiplot}
\vskip -0.1in
\end{figure*}

\subsection{Wiki-Bio Date Results}

\cref{date-multiplot} shows the results for $W\negthinspace$ as \textit{date} vs gender pronoun predictions. The spurious correlations shown in these plots are consistent
with our hypothesized outcome of the selection bias effect, since all of the models were trained on Wikipedia biographical data. Specifically, as \textit{date} increases, women's \textit{access} to resources increases. And thus their representation in Wikipedia increases, inducing the spurious correlation between \textit{date} and \textit{gender}. Due to the nature this collider bias, as female representation goes up, male representation tends to go down\footnote{However, in pre-trained models there exist many selection pressures, one of which appears to cause BERT and RoBERTa to be more likely to predict gender pronouns (as opposed to non-gendered pronouns) for both genders after 2000.}.

\cref{tab:slope-r} shows the slope and Pearson's $r$ correlation coefficient (following \cite{Rudinger18}) for all the plots in this section. These reported coefficients are limited in many ways, including the improbable assumption that the learned relationship between $W\negthinspace$ and \textit{gender} is linear. Further limiting is that we calculate these coefficients against the index of the x-axis, rather than the \textit{date} value on the x-axis. We do this here for consistency with the coming \textit{place} and \textit{subreddit} plots, for which the most convenient quantitative value we can assign to an ordered list of countries or subreddits is their index in the list.

Thus, these coefficients only serve to compare one model’s response to another’s for a given $W\negthinspace$ category. The most noteworthy comparison here is that the correlation coefficients of the pre-trained models are comparable to those of the finetuned models for female pronouns, all above a Pearson's $r$ value of 0.75 (perhaps in part as a trade-off for the male pronoun coefficients).

\subsection{Wiki-Bio Place Results}

The results for $W\negthinspace$ as \textit{place} in \cref{place-multiplot} are similar to those discussed above, although the correlation coefficients appear slightly weaker. As mentioned, reliable comparisons of these coefficients across $W\negthinspace$ variables are limited.
We nonetheless conjecture that the slightly weaker correlations for \textit{place} are perhaps in part due to the subjective nature by which the countries are ordered along the x-axis.

Despite this limitation, we still see comparable slope and correlation coefficients between the finetuned and pre-trained models for the spurious correlation between \textit{place} and \textit{gender} pronouns.

\subsection{Subreddit Results}

A challenge in interpreting the results for $W\negthinspace$ as \textit{subreddit} in \cref{subreddit-multiplot} is that our claim of $W\negthinspace$ as gender-neutral in the `real-word' is dubious for subreddit names, as compared to dates and country names.
As discussed in \cref{sec:x-axis}, we elected against filtering the x-axis to only subreddit names that were more gender-neutral, as this was a subjective process that invited cherry picking. 

To help disambiguate the role of selection bias vs the role of `real-word' gendered terms for $W\negthinspace$, we tested one pre-trained model that was likely exposed to the selection bias effect during pre-training, and one that was likely not exposed. Specifically, RoBERTa should learn a more gendered representation for the otherwise gender-neutral subreddit channel names since the selection bias was likely present during its pre-training, whereas BERT should not.

The bold-faced correlation coefficients in \cref{tab:slope-r} do appear to confirm this hypothesis, with BERT's slope and $r$ coefficients roughly $\frac{1}{5}$ to $\frac{1}{10}$ that of RoBERTa, but the authors note more robust testing is desired to strengthen this argument.

\begin{table*}[]
\caption{Slopes and Pearson's $r$ correlation coefficient for the plots in \cref{all-multiplot} of spurious correlations between \textit{gender} and several otherwise gender-neutral categories for the $W$ variable: \textit{date}, \textit{place} and \textit{subreddit} channel topics. Values in bold font are for the only experiment in which a model was tested against $W$ variables for which the model under test has no selection bias pressures.}
\begin{center}
\begin{small}
\begin{tabular}{ccccccc}
\toprule
\multirow{2}{*}{$W\negthinspace$ Category} & \multirow{2}{*}{Model Type} & \multirow{2}{*}{Model Training Details}        & \multicolumn{2}{c}{Female}                  & \multicolumn{2}{c}{Male}                     \\
                           &                             &                                       				& Slope     & Pearson's r          & Slope     & Pearson's r \\
\midrule        
\midrule                  
\multirow{4}{*}{Date}      
			  & \multirow{2}{*}{finetuned}  
                                                      & WikiBio no Metadata              			&0.558  	&0.929 	&-0.558  	&-0.929 \\
                           &				& WikiBio w Birthdate     				&0.616  	&0.936    	&-0.616 	&-0.936 \\
                           & \multirow{2}{*}{pre-trained} 
                           				& BERT large                             		&0.235  	&0.826	&-0.016 	&-0.116 \\
                           &                             & RoBERTa large                       		&0.149  	&0.759	&-0.055 	&-0.290 \\
\midrule                           
\multirow{4}{*}{Place}      
			  & \multirow{2}{*}{finetuned}  
                                                        & WikiBio no Metadata              			&0.817  	&0.763 	&-0.817  	&-0.763 \\
                           &				& WikiBio w Birthplace     				&0.591  	&0.752    	&-0.591  	&-0.752 \\
                           & \multirow{2}{*}{pre-trained} 
                           				& BERT large                             		&0.381  	&0.476  	&-0.589 	&-0.701 \\
                           &                             & RoBERTa large                       		&0.277  	&0.724	&-0.247 	&-0.525 \\
\midrule
\multirow{4}{*}{Subreddit}      
		  	  & \multirow{2}{*}{finetuned}  
			  				& Reddit TLDR no Metadata 			&0.243     &0.452	&-0.243     &-0.452 \\
                           &                             & Reddit TLDR w Subreddit         		&0.345  	&0.494	&-0.345  	&-0.494 \\
                           
                           & \multirow{2}{*}{pre-trained} 
                           				& BERT large                             		&\bf{0.006}  &\bf{0.049}	&\bf{0.016}  &\bf{0.071} \\
                           &                             & RoBERTa large                       		&0.071  	&0.335	&-0.074 	&-0.300 \\

\bottomrule
\end{tabular}
\label{tab:slope-r}
\end{small}
\end{center}
\end{table*}

\section{Demonstration and Open-Source Code}
The authors would greatly appreciate community feedback that supports or challenges our results. To enable greater access, we have developed a demo of the masked gender task, where users can choose their own input text, as well as the $W\negthinspace$ variable, x-axis values, and the plotted degree of fit, to test spurious correlation to gender in almost any BERT-like model hosted on Hugging Face at \url{https://huggingface.co/spaces/emilylearning/spurious_correlation_evaluation}.

We additionally we will make all code available at \url{https://github.com/2dot71mily/spurious_correlations_ICML_2022}.

\section{Conclusion}

In this paper we have introduced and applied the masked gender task to reveal spurious correlations between gender pronouns and real-world gender-neutral entities like dates and countries, on BERT and RoBERTa large pre-trained models. We showed similar spurious correlations between gender pronouns and subreddit channel names, however as we lacked an objectively gender-neutral list of subreddit names, it is more difficult to disambiguate the role of selection bias vs that of non-gender-neutral topic names. 

The measured correlations between date, country and subreddit-topic vs the probability of a man or woman existing (as a child, adolescent or adult), may be predictive for Wikipedia entries or Subreddit commenters, as $P( Y | X, W\negthickspace, S\negthickspace=\negthickspace1 )$, but will not necessarily generalize to the probabilities of men and women existing in real-world inference domains of $P( Y | X, W)$.

Our results indicate that sentences previously considered as gender-neutral baselines for testing gender bias in LLMs (e.g. input text such as `a woman is walking.' \cite{underspec}), are also vulnerable to spurious correlations.

We explained the role of dataset selection bias in inducing the spurious association between otherwise unconditionally independent entities, such as gender and time, and suggested broad applicability beyond the particular relationships investigated here. 
As mentioned in \cref{recovery}, further assumptions or data can help to mitigate the effects of selection bias, which we hope to apply to LLMs in the future.

\section*{Acknowledgements}
Thank you to the SCIS reviewers for their helpful comments, to Rosanne Liu and Jason Yosinski for their encouragement, to Hugging Face for their open source services, and to Judea Pearl, Elias Bareinboim, Brady Neal and Paul Hünermund for their fantastic online causal inference resources.

\bibliography{GMT}

\begin{thebibliography}{14}
\providecommand{\natexlab}[1]{#1}
\providecommand{\url}[1]{\texttt{#1}}
\expandafter\ifx\csname urlstyle\endcsname\relax
  \providecommand{\doi}[1]{doi: #1}\else
  \providecommand{\doi}{doi: \begingroup \urlstyle{rm}\Url}\fi

\bibitem[Bareinboim and Pearl(2012)]{pearlSelection12}
Elias Bareinboim and Judea Pearl.
\newblock Controlling selection bias in causal inference.
\newblock In Neil~D. Lawrence and Mark Girolami, editors, \emph{Proceedings of
  the Fifteenth International Conference on Artificial Intelligence and
  Statistics}, volume~22 of \emph{Proceedings of Machine Learning Research},
  pages 100--108, La Palma, Canary Islands, 21--23 Apr 2012. PMLR.
\newblock URL \url{https://proceedings.mlr.press/v22/bareinboim12.html}.

\bibitem[Bareinboim and Tian(2015)]{BareinboimTian2015}
Elias Bareinboim and Jin Tian.
\newblock Recovering causal effects from selection bias.
\newblock \emph{Proceedings of the AAAI Conference on Artificial Intelligence},
  29\penalty0 (1), Mar. 2015.
\newblock \doi{10.1609/aaai.v29i1.9679}.
\newblock URL \url{https://ojs.aaai.org/index.php/AAAI/article/view/9679}.

\bibitem[Bareinboim et~al.(2014)Bareinboim, Tian, and Pearl]{selectionrecover}
Elias Bareinboim, Jin Tian, and Judea Pearl.
\newblock Recovering from selection bias in causal and statistical inference.
\newblock \emph{Proceedings of the AAAI Conference on Artificial Intelligence},
  28\penalty0 (1), Jun. 2014.
\newblock URL \url{https://ojs.aaai.org/index.php/AAAI/article/view/9074}.

\bibitem[D'Amour et~al.(2020)D'Amour, Heller, Moldovan, Adlam, Alipanahi,
  Beutel, Chen, Deaton, Eisenstein, Hoffman, Hormozdiari, Houlsby, Hou, Jerfel,
  Karthikesalingam, Lucic, Ma, McLean, Mincu, Mitani, Montanari, Nado,
  Natarajan, Nielson, Osborne, Raman, Ramasamy, Sayres, Schrouff, Seneviratne,
  Sequeira, Suresh, Veitch, Vladymyrov, Wang, Webster, Yadlowsky, Yun, Zhai,
  and Sculley]{underspec}
Alexander D'Amour, Katherine~A. Heller, Dan Moldovan, Ben Adlam, Babak
  Alipanahi, Alex Beutel, Christina Chen, Jonathan Deaton, Jacob Eisenstein,
  Matthew~D. Hoffman, Farhad Hormozdiari, Neil Houlsby, Shaobo Hou, Ghassen
  Jerfel, Alan Karthikesalingam, Mario Lucic, Yi{-}An Ma, Cory~Y. McLean, Diana
  Mincu, Akinori Mitani, Andrea Montanari, Zachary Nado, Vivek Natarajan,
  Christopher Nielson, Thomas~F. Osborne, Rajiv Raman, Kim Ramasamy, Rory
  Sayres, Jessica Schrouff, Martin Seneviratne, Shannon Sequeira, Harini
  Suresh, Victor Veitch, Max Vladymyrov, Xuezhi Wang, Kellie Webster, Steve
  Yadlowsky, Taedong Yun, Xiaohua Zhai, and D.~Sculley.
\newblock Underspecification presents challenges for credibility in modern
  machine learning.
\newblock \emph{CoRR}, abs/2011.03395, 2020.
\newblock URL \url{https://arxiv.org/abs/2011.03395}.

\bibitem[Devlin et~al.(2018)Devlin, Chang, Lee, and Toutanova]{BERT}
Jacob Devlin, Ming{-}Wei Chang, Kenton Lee, and Kristina Toutanova.
\newblock {BERT:} pre-training of deep bidirectional transformers for language
  understanding.
\newblock \emph{CoRR}, abs/1810.04805, 2018.
\newblock URL \url{http://arxiv.org/abs/1810.04805}.

\bibitem[Gokaslan and Cohen(2019)]{OpenWebText}
Aaron Gokaslan and Vanya Cohen.
\newblock Openwebtext corpus, 2019.
\newblock URL \url{http://Skylion007.github.io/OpenWebTextCorpus}.

\bibitem[Heckman(1979)]{heckmanSamplingBias}
James~J. Heckman.
\newblock Sample selection bias as a specification error.
\newblock \emph{Econometrica}, 47\penalty0 (1):\penalty0 153--161, 1979.
\newblock ISSN 00129682, 14680262.
\newblock URL \url{http://www.jstor.org/stable/1912352}.

\bibitem[Knox et~al.(2020)Knox, Lower, and Mummolo]{knox2020}
Dean Knox, Will Lower, and Jonathan Mummolo.
\newblock Administrative records mask racially biased policing.
\newblock \emph{American Political Science Review}, 114\penalty0 (3):\penalty0
  619–637, 2020.
\newblock \doi{10.1017/S0003055420000039}.

\bibitem[Lebret et~al.(2016)Lebret, Grangier, and Auli]{wikibio}
R{\'{e}}mi Lebret, David Grangier, and Michael Auli.
\newblock Generating text from structured data with application to the
  biography domain.
\newblock \emph{CoRR}, abs/1603.07771, 2016.
\newblock URL \url{http://arxiv.org/abs/1603.07771}.

\bibitem[Liu et~al.(2019)Liu, Ott, Goyal, Du, Joshi, Chen, Levy, Lewis,
  Zettlemoyer, and Stoyanov]{RoBERTa}
Yinhan Liu, Myle Ott, Naman Goyal, Jingfei Du, Mandar Joshi, Danqi Chen, Omer
  Levy, Mike Lewis, Luke Zettlemoyer, and Veselin Stoyanov.
\newblock Roberta: {A} robustly optimized {BERT} pretraining approach.
\newblock \emph{CoRR}, abs/1907.11692, 2019.
\newblock URL \url{http://arxiv.org/abs/1907.11692}.

\bibitem[Pearl(2009)]{Pearl09}
Judea Pearl.
\newblock \emph{Causality}.
\newblock Cambridge University Press, Cambridge, UK, 2 edition, 2009.
\newblock ISBN 978-0-521-89560-6.
\newblock \doi{10.1017/CBO9780511803161}.

\bibitem[Rudinger et~al.(2018)Rudinger, Naradowsky, Leonard, and
  Durme]{Rudinger18}
Rachel Rudinger, Jason Naradowsky, Brian Leonard, and Benjamin~Van Durme.
\newblock Gender bias in coreference resolution.
\newblock \emph{CoRR}, abs/1804.09301, 2018.
\newblock URL \url{http://arxiv.org/abs/1804.09301}.

\bibitem[V\"{o}lske et~al.(2017)V\"{o}lske, Potthast, Syed, and Stein]{reddit}
Michael V\"{o}lske, Martin Potthast, Shahbaz Syed, and Benno Stein.
\newblock {TL};{DR}: Mining {R}eddit to learn automatic summarization.
\newblock In \emph{Proceedings of the Workshop on New Frontiers in
  Summarization}, pages 59--63, Copenhagen, Denmark, September 2017.
  Association for Computational Linguistics.
\newblock \doi{10.18653/v1/W17-4508}.
\newblock URL \url{https://www.aclweb.org/anthology/W17-4508}.

\bibitem[Webster et~al.(2020)Webster, Wang, Tenney, Beutel, Pitler, Pavlick,
  Chen, Chi, and Petrov]{websterGenderCorr20}
Kellie Webster, Xuezhi Wang, Ian Tenney, Alex Beutel, Emily Pitler, Ellie
  Pavlick, Jilin Chen, Ed~Chi, and Slav Petrov.
\newblock Measuring and reducing gendered correlations in pre-trained models,
  2020.
\newblock URL \url{https://arxiv.org/abs/2010.06032}.

\end{thebibliography}
\bibliographystyle{icml2022}

\newpage
\appendix
\onecolumn

\section {Explicitly Gendered Words}
See \cref{tab:non-gender-neutral} for list explicitly gendered words that were masked for prediction during both finetuning and at inference time for the masked gender task.
\begin{table}[t]
  \caption{List of explicitly gendered words that are masked out for prediction as part of the masked gender task. These words were largely selected for convenience, as each is a single token in  both the BERT and RoBERTa tokenizer vocabs, for ease of downstream token to word alignment. During finetuning, it is expected that this list will not fully mask gender in every sample, reducing the underspecification of the learning task and the potential learning of gender-neutral spurious associations to gender. At inference time, it is critical that all gendered words are masked, and because the inference input texts are constructed by a heuristic, this is trivial to achieve.}\label{sample-table}
\vskip 0.15in
\begin{center}
\begin{small}
\begin{sc}
\begin{tabular}{cc}
\toprule
    male-variant & female-variant \\	
\midrule
    he & she \\
    him & her \\
    his & her \\
    himself & herself \\
    male & female \\
    man & woman \\
    men & women \\
    husband & wife \\
    father & mother \\
    boyfriend & girlfriend \\
    brother & sister \\
    actor & actress \\
\bottomrule
\end{tabular}
\label{tab:non-gender-neutral}
\end{sc}
\end{small}
\end{center}
\vskip -0.1in
\end{table}

\section{$W\negthinspace$ variable x-axis values}

\subsection{Place Values}
 \label{place-list}
Ordered list of bottom 10 and top 10 World Economic Forum Global Gender Gap ranked countries used for the x-axis in \cref{place-multiplot}, that were taken directly without modification from \url{https://www3.weforum.org/docs/WEF_GGGR_2021.pdf}:

    "Afghanistan",
    "Yemen",
    "Iraq",
    "Pakistan",
    "Syria",
    "Democratic Republic of Congo",
    "Iran",
    "Mali",
    "Chad",
    "Saudi Arabia",
    "Switzerland",
    "Ireland",
    "Lithuania",
    "Rwanda",
    "Namibia",
    "Sweden",
    "New Zealand",
    "Norway",
    "Finland",
    "Iceland"

\subsection{Subreddit Values}
\label{subreddit-list}

Ordered list of subreddits used for the x-axis in \cref{subreddit-multiplot}, that were taken directly without modification from \url{http://bburky.com/subredditgenderratios/} with minimum subreddit size: 400,000.

    "GlobalOffensive",
    "pcmasterrace",
    "nfl",
    "sports",
    "The\_Donald",
    "leagueoflegends",
    "Overwatch",
    "gonewild",
    "Futurology",
    "space",
    "technology",
    "gaming",
    "Jokes",
    "dataisbeautiful",
    "woahdude",
    "askscience",
    "wow",
    "anime",
    "BlackPeopleTwitter",
    "politics",
    "pokemon",
    "worldnews",
    "reddit.com",
    "interestingasfuck",
    "videos",
    "nottheonion",
    "television",
    "science",
    "atheism",
    "movies",
    "gifs",
    "Music",
    "trees",
    "EarthPorn",
    "GetMotivated",
    "pokemongo",
    "news",
    "Fitness",
    "Showerthoughts",
    "OldSchoolCool",
    "explainlikeimfive",
    "todayilearned",
    "gameofthrones",
    "AdviceAnimals",
    "DIY",
    "WTF",
    "IAmA",
    "cringepics",
    "tifu",
    "mildlyinteresting",
    "funny",
    "pics",
    "LifeProTips",
    "creepy",
    "personalfinance",
    "food",
    "AskReddit",
    "books",
    "aww",
    "sex",
    "relationships"

\end{document}